\pdfoutput=1

\documentclass[11pt]{article}

\usepackage[]{acl}

\usepackage{times}
\usepackage{latexsym}

\usepackage[T1]{fontenc}

\usepackage[utf8]{inputenc}

\usepackage{microtype}

\usepackage{graphicx}
\graphicspath{{Figure/}}
\usepackage{amsmath}
\usepackage{amssymb}
\usepackage{amsfonts}
\usepackage{enumitem}
\usepackage{booktabs}
\usepackage{mathrsfs}
\usepackage{multirow}
\usepackage{arydshln}
\usepackage{bm}
\newcommand{\myfont}{\fontsize{10.4pt}{\baselineskip}\selectfont}
\newcommand{\mmyfont}{\fontsize{6.7pt}{\baselineskip}\selectfont}

\title{A Personalized Dialogue Generator with Implicit User Persona Detection}

\author{Itsugun Cho$^1$ \quad Dongyang Wang$^1$\thanks{~~Equal contribution.} \quad Ryota Takahashi$^1$\footnotemark[1] \quad Hiroaki Saito$^1$ \\[3pt]
Keio University, Japan$^1$ \\
{\tt \{choitsugun, wangdongyang, ryota.0226.tokky\}@keio.jp}\\}

\begin{document}
\maketitle
\begin{abstract}
Current works in the generation of personalized dialogue primarily contribute to the agent presenting a consistent personality and driving a more informative response. However, we found that the generated responses from most previous models tend to be self-centered, with little care for the user in the dialogue. Moreover, we consider that human-like conversation is essentially built based on inferring information about the persona of the other party. Motivated by this, we propose a novel personalized dialogue generator by detecting an implicit user persona. Because it is hard to collect a large number of detailed personas for each user, we attempted to model the user's potential persona and its representation from dialogue history, with no external knowledge. The perception and fader variables were conceived using conditional variational inference. The two latent variables simulate the process of people being aware of each other's persona and producing a corresponding expression in conversation. Finally, posterior-discriminated regularization was presented to enhance the training procedure. Empirical studies demonstrate that, compared to state-of-the-art methods, our approach is more concerned with the user's persona and achieves a considerable boost across the evaluations.\let\thefootnote\relax\footnotetext{Submission history: [v1] Fri, 15 Apr 2022.}
\end{abstract}

\begin{figure}
\centering
    \includegraphics[width=6cm]{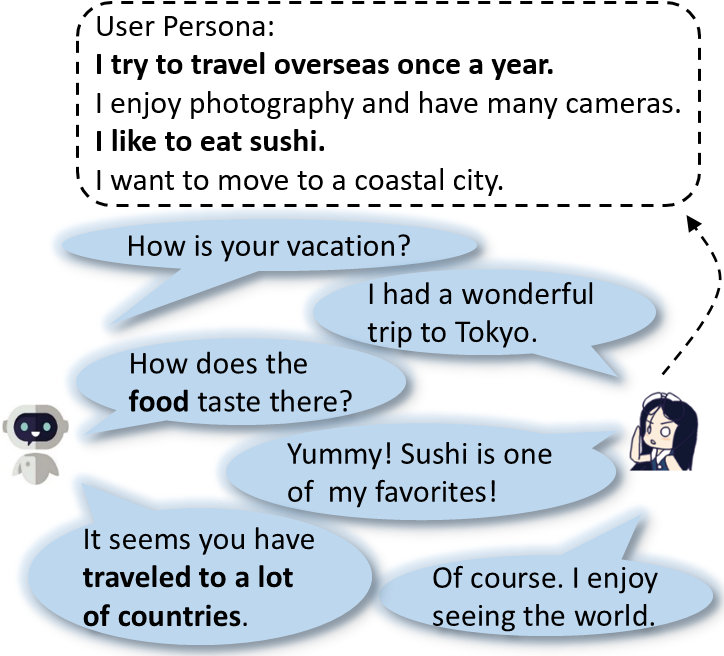}
    \caption{\label{g0} An example of dialogue generation with the implicit persona detection. The incorporated persona and the corresponding user's real persona are in bold.}
\end{figure}

\section{Introduction}
Personalized dialogue modeling is an attractive research topic in deep learning, where studies have explored the possibility of incorporating personal facts into the end-to-end generative framework. The established practice of assigning agents a predefined character improves the engagingness and consistency of open-domain dialogue. However, such models cannot generate distinguishable responses while interacting with different users because they do not take into consideration who the other party is. As \citet{shum2018eliza} pointed out, a good chit-chat bot not only generates interesting responses but also resonates with interlocutors. However, there has been little research conducted into how to make the agent effectively mine a user's persona to generate customized responses.

To this end, this research studied personalized dialogue generation in which we aimed to have the agent recognize the other party's potential persona by exploiting the dialogue itself and output personalized responses conditioned on the different target users. A simple illustration depicting this process is provided in Figure \ref{g0}. Inspired by the impressive effectiveness of conditional variational autoencoders (CVAEs) \cite{sohn2015learning,zhao2017learning} with diverse response modeling, we propose a personalized dialogue generator that detects an implicit user persona using conditional variational inference. Specifically, our model fits the profile descriptions of the other party to a multivariate isotropic Gaussian distribution using a latent variable (perception variable) during training. Because responses from the real-world dialogue are not always persona-related (i.e., persona-sparse issue; \citealp{zheng2020pre}), we also introduce another latent variable (fader variable) to control the weight of persona-related aspects exhibited in the response. During inference, the decoder is designed to acquire the persona features from the perception and fader variables to produce a response that incorporates the user's various potential persona information inferred from the context. Note that the textual profiles are only leveraged during training that is tasked with learning the latent distribution over the user's persona. And during inference, the raw observed data that yields latent variables only includes the context without the explicit persona.

We argue that it is impractical to collect a large quantity of available profiles involved with specific users. Thus, our model has better universality than methods that require providing extra information as generation material. CVAEs have been proved to improve the response diversity at the discourse level (i.e., one-to-many nature; \citealp{zhao2017learning}). Our model achieves “one context to many responses” by sampling and reconstructing with stochasticity for persona distribution and responses, just as we can initiate different chats with a user from aspects of the user's persona. Experimental results on the ConvAI2 dataset demonstrate the superiority of the proposed model over the baselines in both automatic metrics and human evaluations. The interpretability and effectiveness of our approach are clarified in the discussion. The main contributions of this paper can be summarized as: \\
(1) To the best of our knowledge, this is the first attempt to build a user-targeted personalized dialogue agent via conditional variational inference, which not only proposes a new model but also provides insight into manners of latent information mining and representation. \\
(2) A new training scheme is designed to mitigate the disastrous local optimum issue that often occurs in the Bayesian architecture on text generation tasks. Evaluation reveals our scheme yielded better performance than previous strategies. \\
(3) Empirical verification was carried out both quantitatively and qualitatively and confirmed the high levels of convincingness of our model.

\section{Methodology}
\subsection{Problem Scenario}
The task can be formally defined as a dialogue corpus $\mathscr{C} = {(C_i, R_i, P_i)}^{n}_{i=1}$, where $C_i$ refers to a context that includes multiple utterances, with $R_i$ a response and $P_i$ a textual profile containing multiple descriptions of the other party (i.e., the target interlocutor of $R_i$). Our goal is that, by learning the potential dependencies among $P$, $C$, and $R$ from $\mathscr{C}$, one can generate diverse responses $\bar{R} = (\bar{R}_1, \bar{R}_2, ..., \bar{R}_m)$ for a new context $\bar{C}$. $\bar{R}$ is expected to be relevant to the other party's real persona, which means the mutual information should be maximized as much as possible. Moreover, in cases where $\bar{C}$ is the persona-sparse context, $\bar{R}$ should mainly cohere with the context.

\begin{figure}
\centering
    \includegraphics[width=6cm]{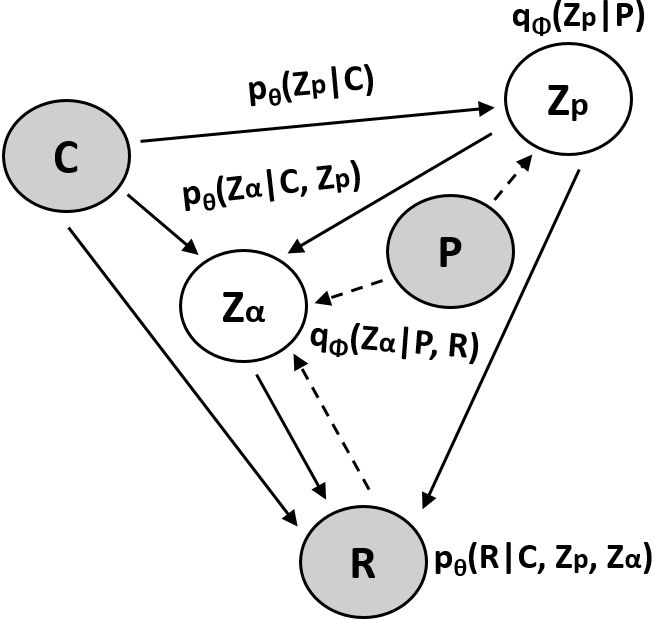}
    \caption{\label{g1} The solid lines are conditional dependencies and dashed lines denote variational approximation. The profile $P$, context $C$, and response $R$ are observed data. The variational parameters $\phi$ are learned jointly with the conditional parameters $\theta$.}
\end{figure}

\begin{figure*}[t]
\centering
    \includegraphics[clip,width=16cm]{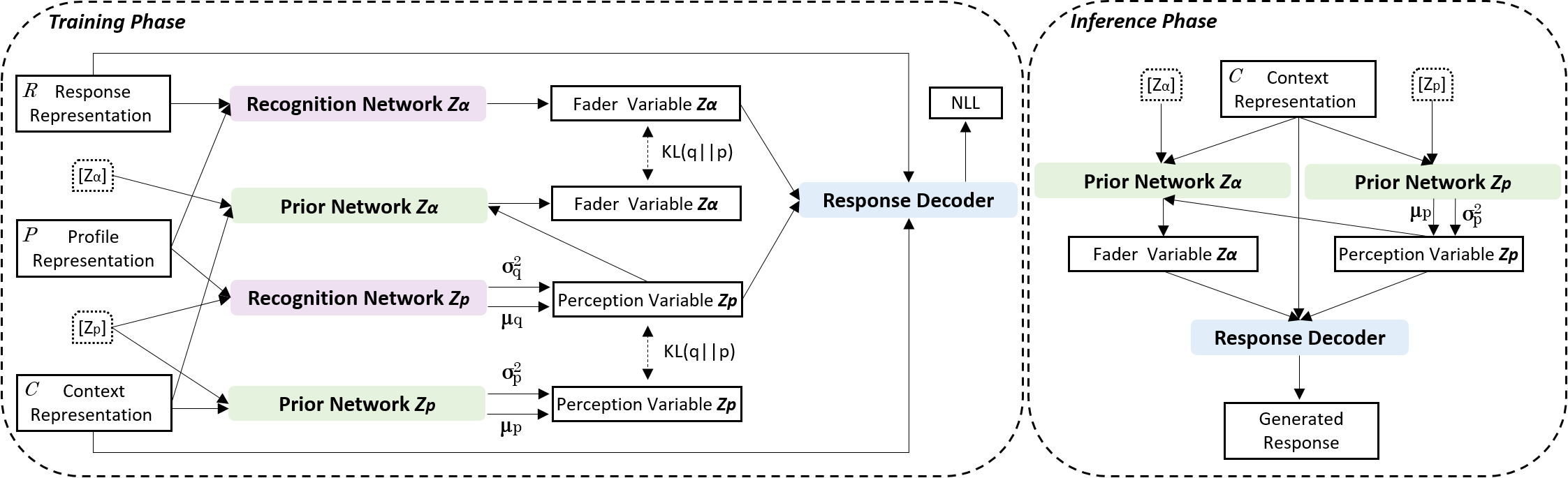}
    \caption{\label{g2} Illustration of the model architecture. The two prior networks share parameters.}
\end{figure*}

\subsection{Overview}
As described in the introduction, our approach incorporates a pair of latent variables utilized for bridging the potential dependencies among $P$, $C$, and $R$. Perception variable $Z_p$ is adopted to capture the latent distribution over $P$ that constructs a connection between $C$ and $R$ by the user's implicit persona. Fader variable $Z_\alpha$ is adopted to indicate how much persona information in $Z_p$ is carried by $R$ under $C$. Figure \ref{g1} gives the directed graphical model of our approach. The conditional distribution over the above variables can be factorized as $p(R, Z_p, Z_\alpha|C) = p(R|C, Z_p, Z_\alpha)p(Z_\alpha|C, Z_p)p(Z_p|C)$. Our objective is to represent it with deep neural networks, where we denote $p_\theta(R|C, Z_p, Z_\alpha)$ as a response decoder and $p_\theta(Z_p|C)$ and $p_\theta(Z_\alpha|C, Z_p)$ as the prior networks. $p(R, Z_p, Z_\alpha|C)$ depicts a process that is from the prior networks to draw out implicit persona and its representation from $C$, prompting the response decoder to restore $R$ under the information only sourced in $C$. Thereby, we would maximize the conditional likelihood \myfont$p_\theta(R|C) = \iint p_\theta(R|C, Z_p, Z_\alpha)p_\theta(Z_\alpha|C, Z_p)p_\theta(Z_p|C)dZ_p dZ_\alpha$\normalsize.

However the marginalization over $Z_p$ and $Z_\alpha$ are intractable integrals (i.e., a context theoretically corresponds to a continuous user persona space). Hence, our model is trained with the stochastic gradient variational Bayes (SGVB) framework \cite{kingma2013auto} by maximizing the variational lower bound. According to the above definition of the perception and fader variables, we refer to variational distribution $q_\phi(Z_p|P)$ and $q_\phi(Z_\alpha|P, R)$ as the recognition networks to approximate the true posterior $p(Z_p|C, R)\propto p(R|C, Z_p)p(Z_p|C)$ and $p(Z_\alpha|C, R, Z_p)\propto p(R|C, Z_p, Z_\alpha)p(Z_\alpha|C, Z_p)p(Z_p|C)$,\ respectively. The evidence lower bound (ELBO) of our approach can be deduced as follows:
\begin{equation}
\begin{split}
&\mathcal{L}(\theta, \phi; P, C, R) = \\
&-KL(q_\phi(Z_p|P)||p_\theta(Z_p|C)) \\
&-KL(q_\phi(Z_\alpha|P, R)||p_\theta(Z_\alpha|C, Z_p)) \\
&+\mathbb{E}_{q_\phi(Z_p|P);q_\phi(Z_\alpha|P, R)}[{\rm log} p_\theta(R|C, Z_p, Z_\alpha)]
\end{split}
\end{equation}
where $KL(\cdot||\cdot)$ denotes the KL divergence. Details about the derivation are provided in Appendix A.

\subsection{Model Details}
Figure \ref{g2} shows the architecture of our model. We define the input representation as follows: \\
(1) The input embedding of each token is the sum of corresponding word embedding and position embedding. To differentiate the user character in dialogue history, we add role embedding into the utterances generated by the other party. With minor exploitation of notation, we also use $P$, $C$, and $R$ to denote input representations in the following. \\
(2) The different utterances in context or different descriptions in the profile are separated by the special token [SEP]. The beginning and end of the context or profile are appended with the special tokens [BOS] and [EOS], respectively. \\
(3) The special token of the perception variable and the fader variable are denoted as [$Z_p$] and [$Z_\alpha$], respectively. For the special token of the latent variable, the position embedding is set to empty.

We hypothesize the perception variable follows multivariate Gaussian distribution with a diagonal covariance matrix. The input representations $concat([Z_p], C)$ and $concat([Z_p], P)$ are fed to the prior network $p_\theta(Z_p|C)\backsim \mathcal{N}(\bm{\mu}_p, \bm{\sigma}^2_p\textbf{I})$ and the recognition network $q_\phi(Z_p|P)\backsim \mathcal{N}(\bm{\mu}_q, \bm{\sigma}^2_q\textbf{I})$, respectively, where $concat(\cdot\,,\cdot)$ denotes concatenation. Both networks are three-layer transformer encoders \cite{vaswani2017attention} with a two-layer fully connected network. The means $\bm{\mu}_p$, $\bm{\mu}_q$ and variances $\bm{\sigma}^2_p$, $\bm{\sigma}^2_q$ are derived as follows:
\begin{equation}
\begin{bmatrix} \bm{\mu}_{p} \\{\rm log}(\bm{\sigma}_{p}^{2}) \end{bmatrix} = \bm{W}_{p}\ \bm{h}_{[Z_p]} + \bm{b}_{p}
\end{equation}
\begin{equation}
\begin{bmatrix} \bm{\mu}_{q} \\{\rm log}(\bm{\sigma}_{q}^{2}) \end{bmatrix} = \bm{W}_{q}\ \bm{h}_{[Z_p]} + \bm{b}_{q}
\end{equation}
where $\bm{h}_{[Z_p]}\in\mathbb {R}^D$ is the final hidden state of [$Z_p$] from the transformer encoder, and $\bm{W}_p\in\mathbb {R}^{K\times D}$, $\bm{W}_q\in\mathbb {R}^{K\times D}$, and $\bm{b}_p\in\mathbb {R}^K$, $\bm{b}_q\in\mathbb {R}^K$ denote the weight matrices of the fully connected network. We obtain samples of the perception variable from $\mathcal{N}(\bm{\mu}_q, \bm{\sigma}^2_q \textbf{I})$ during training or $\mathcal{N}(\bm{\mu}_p, \bm{\sigma}^2_p \textbf{I})$ during inference. As sampling is not differentiable, the re-parametrization trick \cite{kingma2013auto} is employed for effective training.

\begin{figure}
\centering
    \includegraphics[width=6cm]{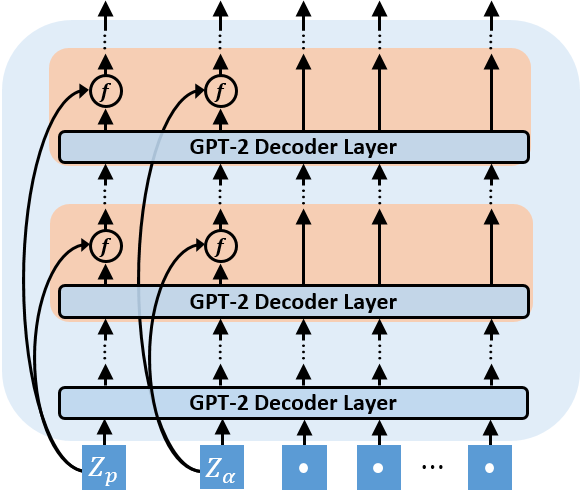}
    \caption{\label{g3} Over each multiple transformer layer (four layers in our experiments), a weighted sum \emph{\textbf{f}} is operated between the hidden state of latent variables and the original latent variables. Both inputs of operation \emph{\textbf{f}} are weighted 0.5 in our implementation.}
\end{figure}

The input representation $concat(Z_p, [Z_\alpha], C)$ is fed to the prior network $p_\theta(Z_\alpha|C, Z_p)$, which is a three-layer transformer encoder, and the final hidden state of [$Z_\alpha$] is specified as a fader variable. The recognition network $q_\phi(Z_\alpha|P, R)$ without parameters concerns a similarity function of $(P_i, R_i)^{n}_{i=1}$ pairs. We obtain the fader variable from $q_\phi(Z_\alpha|P, R)$ during training or $p_\theta(Z_\alpha|C, Z_p)$ during inference. The response decoder $p_\theta(R|C, Z_p, Z_\alpha)$ is built by a GPT-2 pre-trained language model \cite{radford2019language}. The input representations $concat(Z_p, Z_\alpha, C, R)$ or $concat(Z_p, Z_\alpha, C)$ are fed to the response decoder during training or inference, respectively. Note that we put $Z_p$ and $Z_\alpha$ before $C$, $R$, or $C$ to form the input representations due to the autoregressive property of GPT-2. To facilitate the backpropagation of the perception and fader variables, and also to enhance the effect of these variational signals on generation in decoding, we considered an injection scheme that is illustrated in Figure \ref{g3}.

\subsection{Posterior-Discriminated Regularization}
Training the text data with VAEs / CVAEs often falls into a trivial local optimum where the decoder learns to ignore the latent variable, causing the approximate posterior to mimic the prior. This phenomenon is referred to as “posterior collapse.” The state-of-the-art solutions include re-weighting the KL term (KL annealing, cyclic annealing; \citealp{bowman2016generating,fu2019cyclical}), introducing a neural network to calculate bag-of-word (BOW) loss \cite{zhao2017learning}, and modifying the training procedure (aggressive training; \citealp{he2019lagging}).

Ideally, if the approximate posterior $q_\phi(Z|X)\backsim \mathcal{N}(\bm{\mu}_q, \bm{\sigma}^2_q \textbf{I})$ (i.e., $q_\phi(Z_p|P)$ in our experiment) is perfect, $Z$ is a non-trivial latent representation of input $X$, whereby we suppose that $Z$ should be especially dissimilar for various posterior inputs. We designed a scheme augmenting the distinction of conditional posteriors that forces the decoder to reconstruct results from the latent variable whose features vary notably. We consider training an auxiliary cost with minimizing the following.
\myfont
\begin{equation}
\begin{split}
&\mathcal{L}_{Po\raisebox{0mm}{-}di} = \\
&\sum^n_{i=1} (Min[KL(q_\phi(Z_i|X_i)||q_\phi(\bar{Z_i}|\bar{X_i})) - \lambda, 0])^2
\end{split}
\end{equation}
\normalsize
where $X_i$ denotes the $i$-th training data, and $\bar{X_i}$ refers to the input other than $X_i$. The distinction objective $\lambda$ drives up KL divergence between the posteriors over different inputs. In our implementation, this computation is dealt with as mini-batch processing which the data is random sampling without replacement. The auxiliary cost can be added to ELBO to form the final loss function.
\begin{equation}
\begin{split}
&\mathcal{L}^{\prime}(\theta, \phi; P, C, R) = \\
&\mathcal{L}(\theta, \phi; P, C, R) + \mathcal{L}_{Po\raisebox{0mm}{-}di}
\end{split}
\end{equation}
Despite being conceptually simple, the benefit of this idea is that it is task-independent and easy to train without introducing new model components. $\mathcal{L}_{Po\raisebox{0mm}{-}di}$ achieves better performance, as we will detail in Section 4.3 by comparing the above methods.

\section{Experiments}
\subsection{Corpus}
We evaluated our approach on the ConvAI2 benchmark dataset, which is an extended version with a new hidden testing set of the PERSONA-CHAT dataset \cite{zhang2018personalizing}. The dialogues were collected from crowd-workers who were asked to act as two interlocutors having a conversation to get to know each other. The persona of both interlocutors is explicitly described using several profile sentences. This dataset contains 17,878 / 1,000 multi-turn dialogues conditioned on 1,155 / 100 profiles for train / dev, each profile consisting of at least five descriptions. Because the testing set is hidden, we used the validation set as the testing set in our experiments and randomly sampled 500 dialogues from the training set for the validation. To suit our goals, we removed some self-centered utterances that only scratched the surface.

\subsection{Baselines}
The following five state-of-the-art generative baseline methods were considered in our experiments. \\
\textbf{HRED} is a persona-free dialogue model built by hierarchical RNN, proposed in \citet{serban2016building}. This model is one of the traditional seq2seq architectures widely applied for comparison. \\
\textbf{CVAE} is a persona-free dialogue model utilizing a conditional variational autoencoder to learn a latent distribution over conversational factors. This model was proposed by \citet{zhao2017learning}. \\
\textbf{TTransfo} is a GPT-based personalized dialogue model with multi-task learning proposed by \citet{wolf2019transfertransfo}. This model obtained state-of-the-art performance on automatic metrics in the Second Conversational Intelligence Challenge. \\
\textbf{\bm{$P^2$} BOT} is a GPT-based personalized dialogue model with the reinforce algorithm proposed by \citet{liu2020you}. This is the latest state-of-the-art model for dialogue generation on persona-chat. \\
\textbf{DialoGPT} is a pre-trained dialogue model proposed by \citet{zhang2020dialogpt}. This model is based on GPT-2 using the Reddit comments dataset. We compared ours to the version of model size 345M, which had the best result reported in the paper.

\subsection{Implementation}
Our implementation was based on the PyTorch \cite{paszke2019pytorch} and HuggingFace libraries \cite{wolf2019huggingface}. The response decoder GPT-2 was set to 16 heads, 24 layers, 1024 dimensional hidden state, and with 345M parameters. All input representation refers to the embedding tables of GPT-2, and the embedding size was the same setting as the size of latent variables, which was fixed at 1024. The distinction objective $\lambda$ was set to 0.15. The Adam algorithm \cite{kingma2015adam} was utilized for optimization with a learning rate of 2.6e-5, and a warmup step of 3000. Responses were generated by nucleus filtering \cite{holtzman2019curious} where top-k and top-p were set to 4 and 0.8, respectively. BPE algorithm \cite{sennrich2016neural} was used for word tokenization, the token vocabularies of GPT2, with a size of 50,257, were shared by the prior and posterior networks.

\begin{table*}
\centering
\begin{tabular}{ccccccc}
\toprule
\textbf{Model} & \textbf{PPL} & \textbf{Distinct-1 / 2} &\textbf{P.Distance} &\textbf{Coherence} &\textbf{Engagingness} &\textbf{P.Relevancy}\\
\toprule
{HRED}        &21.095 &0.078 / 0.225 &0.246 &0.657 &0.703 &0.670\\
{CVAE}        &19.501 &0.116 / 0.405 &0.258 &0.557 &0.673 &0.523\\
{TTransfo}    &18.011 &0.142 / 0.400 &0.329 &0.840 &0.703 &0.877\\
{DialoGPT}    &\textbf{14.966} &0.139 / 0.417 &0.359 &1.037 &0.883 &0.900\\
{$P^2$ BOT}   &16.620 &0.083 / 0.268 &0.370 &0.953 &0.920 &0.873\\
{Ours}        &15.671 &\textbf{0.167} / \textbf{0.538} &\textbf{0.401} &\textbf{1.177} &\textbf{1.203} &\textbf{1.207}\\
\bottomrule
\end{tabular}
\caption{\label{t1} Evaluation results on the ConvAI2 dialogue corpus, the best score in each metric are in bold. For our model and CVAE, the latent variables were sampled N times to generate N responses, and the final evaluation scores were acquired by average (N = 3 in our experiments); 50 dialogues were randomly sampled from the testing set for human evaluation. The statistical test showed the differences are significant with p-value $<$ 0.05.}
\end{table*}

\subsection{Evaluation}
\subsubsection{Automatic Metrics}
We followed previous work and employed \textbf{Perplexity (PPL)} \cite{sutskever2014sequence} and \textbf{Distinct} \cite{li2015diversity}. The PPL measures the negative log-likelihood of the ground-truth sequence output by the model. A lower PPL generally indicates that the learned language model is more human-like. The Distinct is calculated as the number of distinct unigrams and bigrams divided by the total number of generated words. This metric assesses the degrees of word-level diversity for generated responses.

Furthermore, we propose a new metric to estimate the level of the correlation of generated response and the user's persona, which is named \textbf{P.Distance (Persona Distance)}. For word embedding trained under the language model, the distance between vectors in the respective space is proportional to the relative co-occurrence of words they represent. Therefore, we employed the pre-trained Google News (300D)\footnote{https://code.google.com/archive/p/word2vec/} word2vec to measure the closeness between the response and corresponding profile in the vector space. We removed stop words for the profile and the generated response, then extracted the keywords of each response-profile pair by word frequency of the training set. For the $i$-th profile keyword embedding $\bm{p}_i$, we can make the similarity matrix as follows:
\myfont
\begin{equation}
\begin{split}
&\bm{M}_i = \\
&[Sim(\bm{p}_i, \bm{r}_1), Sim(\bm{p}_i, \bm{r}_2), …, Sim(\bm{p}_i, \bm{r}_n)]
\end{split}
\end{equation}
\normalsize
where $Sim(\cdot\,,\cdot)$ is a cosine similarity function, and $\bm{r}_i$ is the embedding of the $i$-th response keyword. The P.Distance can be calculated as follows:
\myfont
\begin{equation}
\begin{split}
&P.Distance = \\
&Ave(Max(\bm{M}_1), Max(\bm{M}_2), …, Max(\bm{M}_n))
\end{split}
\end{equation}
\normalsize

\subsubsection{Human Metrics}
We engaged six native speakers\footnote{All the annotators are graduate students recruited from the internet whose are not relevant to this study.} to annotate the quality of generated responses based on the following criteria. The scale of these metrics is [0, 1, 2], and for each dialogue, the generated responses by all models were order shuffled in the evaluation. \\
\textbf{Coherence} measures whether the response is consistent with the context. \emph{Score 0}: The response is not related to the context. \emph{Score 1}: The response mentions something related to the context but is not coherent. \emph{Score 2}: The response is coherent with the context and not generic. \\
\textbf{Engagingness} assesses how well the response endeavors to continue the dialogue. \emph{Score 0}: The response is generic or poor quality, which makes it difficult to continue the dialogue. \emph{Score 1}: The response is boring, but it is still acceptable to continue the dialogue. \emph{Score 2}: The response is interesting and the dialogue can be developed. \\
\textbf{P.Relevancy (Persona Relevancy)} estimates the degree of a response being relevant to the other party's persona. And the persona of the other party is required to be inferred from the context. \emph{Score 0}: What the response mentions is irrelevant to the other party's persona. \emph{Score 1}: The response involves a question to the other party and is not generic. \emph{Score 2}: What the response mentions is related to the persona of the other party.

\subsubsection{Results}
Table \ref{t1} shows the evaluation results. We can see that, compared to baseline, our approach was superior in all metrics except PPL. Nonetheless, this metric also achieves highly competitive performance. CVAE gains a higher Distinct score than HRED and $P^2$ BOT, which could be attributed to the variational autoencoder catching discourse-level diversity. Our model surpassed CVAE on Distinct and P.Distance, which suggests that the implicit persona modeling can better reflect the specific user's persona, creating more informative responses. The comparison with DialoGPT can be seen as an ablation study since our model would degenerate into a GPT-2 with removal of the latent variables. It can be observed that ours outperformed DialoGPT overall, which reveals that the proposed latent variables are beneficial for generating more user-related and diversified responses. Ours is slightly inferior on PPL, which is to be expected due to stochasticity for the language model brought by the latent variables.

On the other side, the personalized dialogue models TTransfo and $P^2$ BOT received an undesirable P.Relevancy. And $P^2$ BOT had only a slightly improved Engagingness compared to DialoGPT, which indicates that even with a specific personality, responses that lack consideration for the interlocutor may limit the attraction for people to continue the exchange. By contrast, ours obtained meaningful advances in Engagingness and P.Relevancy, which demonstrates that responses relevant to the other party's persona can motivate the user to participate actively in conversation. When it comes to Coherence, both HRED and CVAE attained lower scores compared to all other large-scale transformer-based models. This is not surprising because pre-trained language models have proved to have better language understanding capability than traditional RNNs. The Fleiss' kappa \cite{fleiss1971measuring} score with human judges was around 0.33, which can be regarded as “fair agreement.”

\section{Discussion}
\subsection{Analyzing Latent Variables}
\begin{figure}[t]
\centering
    \includegraphics[width=6cm]{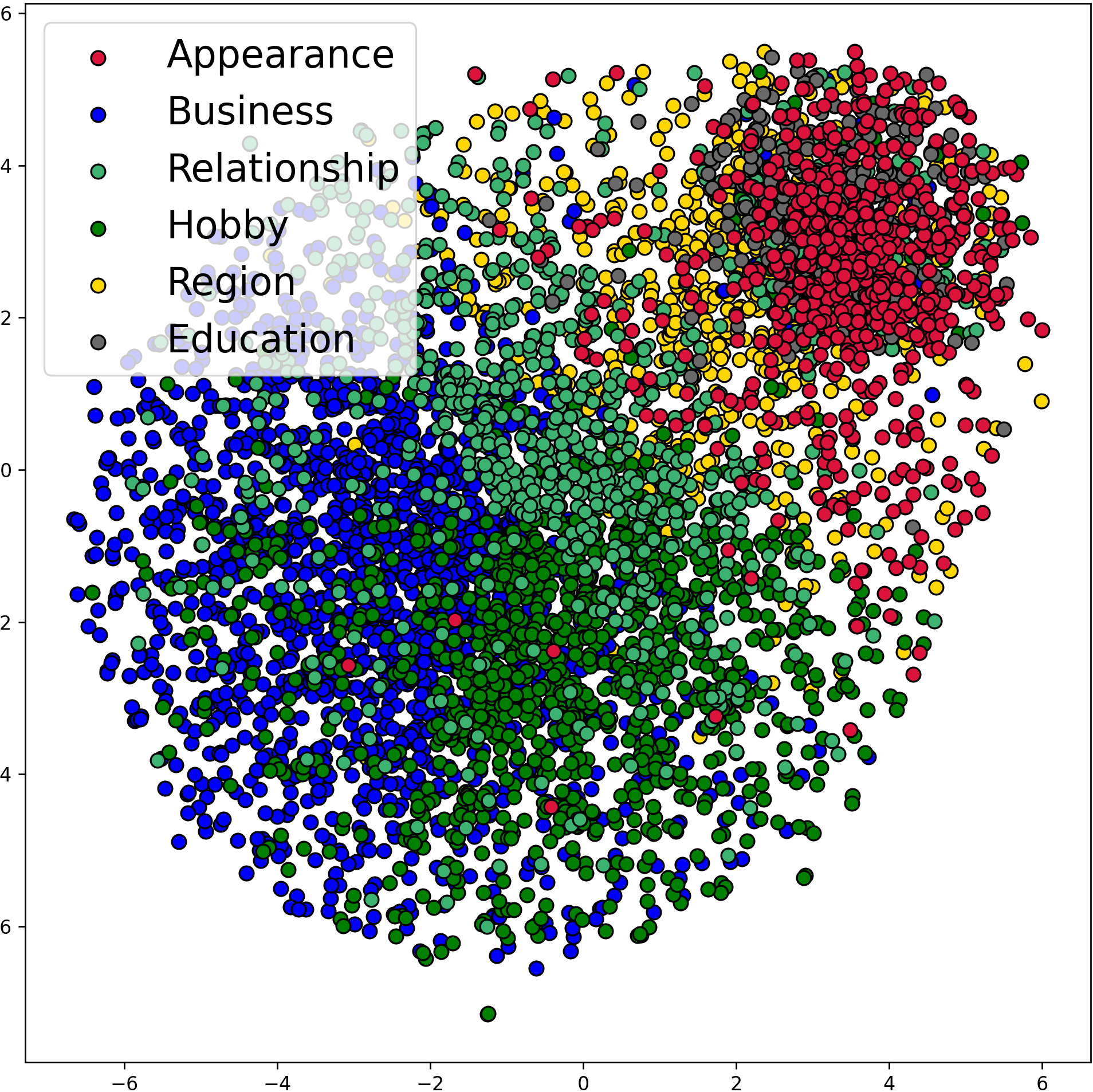}
    \caption{\label{g4} The visualization of perception variable.}
\end{figure}

One assumption is that individual persona features could be classified in latent space. Additionally, previous research \cite{zhao2017learning} has identified that the posterior network can grasp the clustering of high-dimensional discrete samples. Thus, we wanted to check if the perception variable can be learned in the explainable collections. All profiles in the training set were classified into six pre-defined categories by employing a pre-trained zero-shot classifier \cite{lewis2020bart}. The classifier calculates the probabilities of category attribution in the manner of building profiles and categories into premise-hypothesis pairs \cite{yin2019benchmarking}. Figure \ref{g4} visualizes the posterior perception variables in 2D space using t-SNE \cite{van2008visualizing}. We discovered that the latent space learned by $Z_p$ is correlated with the profile categories. Recall that perception variable is devised to refine the user's implicit persona, and this result is in line with our initial conception.

\begin{figure}
\centering
    \includegraphics[width=7.7cm]{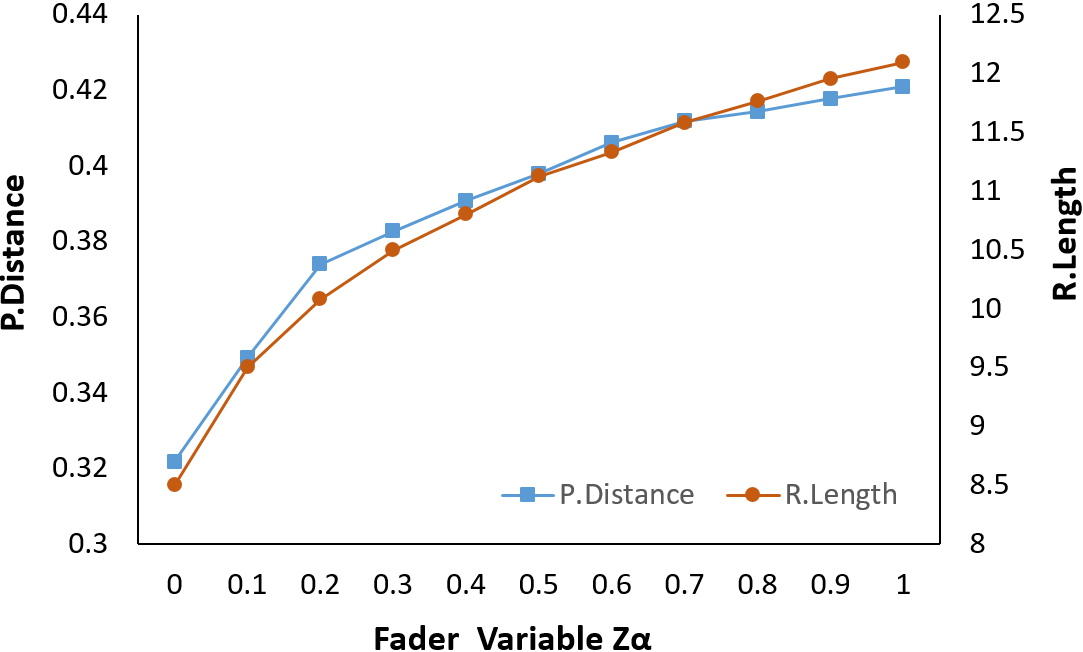}
    \caption{\label{g5} Controllability analysis for fader variable. The increment is set as 0.1.}
\end{figure}

Then we studied the impact of the fader variable in modeling response generation. Because the fader variable aims to control the representation of implicit persona, we verified its effect on the generation by sliding the value. Specifically, we gradually boosted $Z_\alpha$ from 0 to 1 instead of the prior network $p_\theta(Z_\alpha|C, Z_p)$ during inference. The proximity between the generated response and ground-truth profile was computed by P.Distance. Figure \ref{g5} reveals the test result that the proximity had an inverse correlation with increasing value of the fader variable. Meanwhile, the length of generated response (the number of generated tokens in a response) showed an increasing tendency due to the fader variable controlling the amount of persona information the model was attempting to represent.

\begin{table}[t]
\centering
\begin{tabular}{c|l}\toprule
\multicolumn{2}{l}{\textbf{User Persona:}} \\
\multicolumn{2}{l}{I like sports, especially basketball. \quad I am tall.} \\
\multicolumn{2}{l}{I like listening to music. \quad I am a student.} \\
\bottomrule
\multirow{4}{*}{\rotatebox[origin=c]{90}{1st Turn}} & U: \,Hi, I like listening to music. \\
& $r_1$: Cool, my favorite \textbf{musician} is Todd Lee. \\
& $r_2$: Who is your favorite \textbf{singer}?  \\ 
& $r_3$: That's cool do you have pets? \\ \hline
\multirow{5}{*}{\rotatebox[origin=c]{90}{2nd Turn}} & U: \,I also enjoy playing basketball. \\
& $r_1$: That is fun. What \textbf{position} do you play? \\
& $r_2$: What \textbf{team} do you like the least?  \\ 
& $r_3$: Oh, that's interesting. \\
& \quad\,\, How long have you played basketball?  \\ \hline
\multirow{6}{*}{\rotatebox[origin=c]{90}{3rd Turn}} & U: \,I play center. \\
& $r_1$: Oh, that's awesome. How \textbf{tall} are you? \\
& $r_2$: Ah, nice. \\
& \quad\,\, Do you play for a \textbf{team} or just for fun?  \\ 
& $r_3$: Oh, that is cool. \\ 
& \quad\,\, I am starting to get into the \textbf{NBA}. \\ \hline
\multirow{6}{*}{\rotatebox[origin=c]{90}{4th Turn}} & U: \,I am 6 foot 3. \\
& \quad\, I often play with my roommates. \\
& $r_1$: Wow, that's tall. Are you in \textbf{school}? \\
& $r_2$: I bet you can play \textbf{baseball} too.  \\ 
& $r_3$: I am 6 feet 1 inches tall. \\
& \quad\,\, Do you have siblings? \\
\bottomrule
\end{tabular}
\caption{\label{t2} The involved personas in responses are in bold.}
\end{table}

\subsection{Case Study}
For a more comprehensive qualitative analysis, we set up a hypothetical persona to interact with our model. Table \ref{t2} presents the diverse responses generated from our model in four consecutive turns. The responses were sampled randomly three times per turn, and utterances from the user follow by $r_1$ of the previous turn. As observed, most of the responses not only covered the user's persona but were also consistent with the context. Responses in each turn are significantly dissimilar, which suggests our model can perform various expressions by the proposed latent variables. That is in line with the “one-to-many” nature mentioned in the introduction. Further, we detected that our model was more likely to raise questions about the other party and seldom generated generic responses. Naturally, putting forth a question could contribute to keeping the conversation going. And such responses also have a high probability of hitting an aspect of the user's persona. That explains why our approach obtained a remarkable score on Engagingness and P.Relevancy in the human evaluation.

\subsection{Probing \bm{$\mathcal{L}_{Po\raisebox{0mm}{-}di}$}}
\begin{figure}[t]
\centering
    \includegraphics[width=7.7cm]{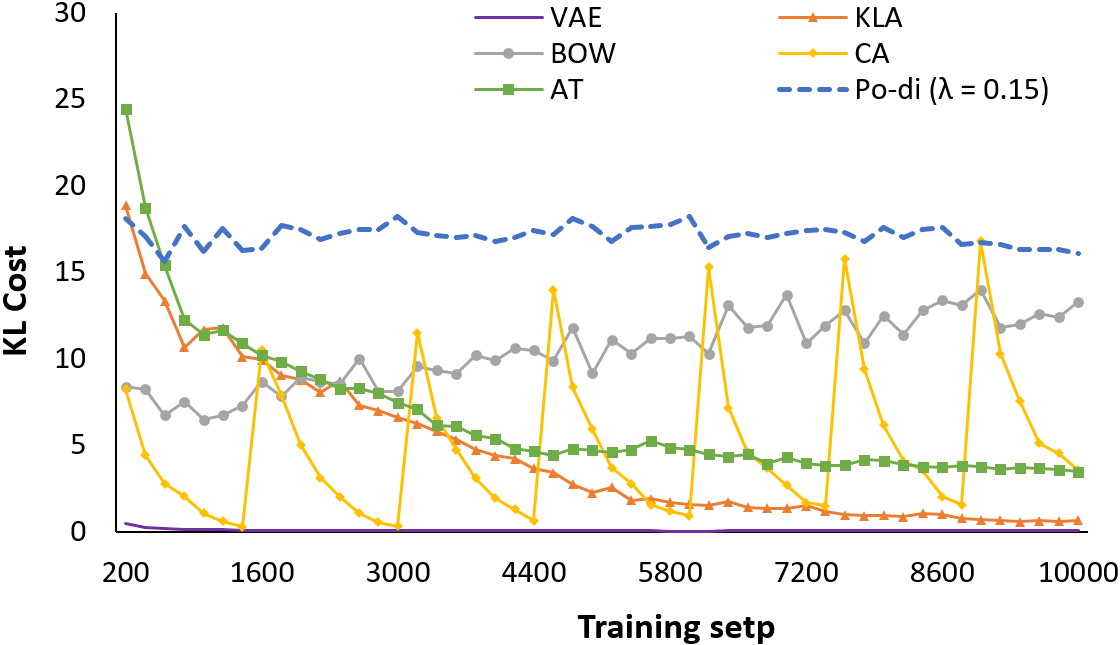}
    \caption{\label{g6} The convergence of KL costs during training.}
\end{figure}

The efficacy of $\mathcal{L}_{Po\raisebox{0mm}{-}di}$ in helping to alleviate “posterior collapse” was assessed by a comparative trial. We carried out the language modeling task on Penn Treebank \cite{marcinkiewicz1994building} utilizing the VAE constructed by seq2seq architecture based on GRU. We set the KL weight of KL annealing (KLA) to increase linearly from 0 to 1 in the first 5000 steps. Table \ref{t3} reports PPL, the number of active units (AU) \cite{burda2016importance}, and KL cost for six kinds of training techniques on the testing set. We varied the distinction objective $\lambda$ and report four settings between 0.12 and 0.21. In our experiments, the settings in this range obtained a sounder balance between PPL and KL cost. We can see that $\mathcal{L}_{Po\raisebox{0mm}{-}di}$ reconstructed the language model with lower perplexity while converging to a small but meaningful KL cost. $\mathcal{L}_{Po\raisebox{0mm}{-}di}$ retained more active units than others, which indicates a richer latent representation can be acquired by “pulling apart” the KL divergence between the different posteriors. Figure \ref{g6} presents the evolution of the KL cost during training. Compared to VAE without any strategies, the model with KLA can prevent the KL cost crashes at the beginning of training, but the effect is diminished by degrees after the KL weight climbs to 1. Although this problem is fixed by cyclic annealing (CA) and aggressive training (AT), they still have slightly poor performance on PPL. VAE with BOW gained comparable performance to ours, whereas $\mathcal{L}_{Po\raisebox{0mm}{-}di}$ without introducing any supplemental neural network still mitigated “posterior collapse.”

\begin{table}[t]
\centering
\begin{tabular}{cccc}
\toprule
\textbf{Method} &\textbf{PPL} &\textbf{AU} &\textbf{KL cost} \\
\toprule
{Standard VAE}&58.391 &0 &0.049 \\
\hdashline
{+ KLA}&53.564 &3 &2.508 \\
{+ CA} &51.547 &2 &3.767 \\
{+ AT} &50.000 &6 &5.320 \\
{+ BOW}&48.637 &11 &11.165 \\
{+ $\mathcal{L}_{Po\raisebox{0mm}{-}di}$ $(\lambda=0.12)$} &48.467 &13 &14.727 \\
{+ $\mathcal{L}_{Po\raisebox{0mm}{-}di}$ $(\lambda=0.15)$} &47.674 &\textbf{14} &16.096 \\
{+ $\mathcal{L}_{Po\raisebox{0mm}{-}di}$ $(\lambda=0.18)$} &\textbf{45.925} &\textbf{14} &22.047 \\
{+ $\mathcal{L}_{Po\raisebox{0mm}{-}di}$ $(\lambda=0.21)$} &46.883 &\textbf{14} &25.126 \\
\bottomrule
\end{tabular}
\caption{\label{t3} Automatic results for different methods.}
\end{table}

\section{Related Work}
\subsection{Variational Autoencoders (VAEs)}
The VAEs \cite{kingma2013auto,rezende2014stochastic} were proposed for image generation and applied by \citet{bowman2016generating} for natural language generation. Then, the CVAEs \cite{yan2016attribute2image,sohn2015learning} were proposed to enable more controllable generation that conditioned certain attributes. \citet{zhao2017learning} adopted the CVAE for the task of multi-turn dialogue modeling, which learns a distribution over dialogue acts to capture discourse-level variations. The above models achieve various generations by drawing latent variables from the learned distribution.

\subsection{Personalized Dialogue Models}
Recently, there has been much research exploring different approaches to the task of personalized dialogue generation \cite{yang2020multitask,song2020generating,zheng2020pre,wu2020guiding,xu2021diverse}. $P^2$ BOT \cite{liu2020you} and TTransfo \cite{wolf2019transfertransfo} are recognized state-of-the-art baselines on persona-chat. $P^2$ BOT proposes a transmitter-receiver and mutual persona perception framework that fuses supervised training and self-play fine-tuning for enhancing the quality of personalized dialogue generation. TTransfo combines transfer learning and the Transformer model, and fine-tuning is performed on the pre-trained model by optimizing the multi-task objective function to improve the fluency of personalized responses. The aforementioned approaches involve conditioning responses on the additional agent's persona. Instead, the variational method allows us to be flexible in handling the effects of conditions (i.e., context) and is independent of external knowledge. Our method further integrates the details of the user into the inference process.

\section{Conclusion and Future Work}
This paper presented a new implicit persona detection generator to achieve a user-personalized response. We establish the persona exploration and dialogue generation in a unified framework that supplies a way of leveraging the potential facts in dialogue. Experiments on a large public dataset demonstrated that our approach had superior performance in producing user-specific responses. Humans typically continue exchanges by drilling into the content of the conversation. From this perspective, the PersonalDialog dataset \cite{zheng19personalized} may be more appropriate for our approach. In the future, we plan to use this dataset to study if the inference of implicit personas can be strengthened. We would also conduct further experiments to examine whether there is an interpretable association between the prior network and the recognition network in terms of what they have learned. Eventually, we plan to perfect $\mathcal{L}_{Po\raisebox{0mm}{-}di}$ by having weights flexibly regulate the KL divergence of posteriors.

\bibliography{anthology,custom}
\bibliographystyle{acl_natbib}

\appendix
\section{Appendix}
\subsection*{Derivation of ELBO}
The conditional likelihood can be written as follows by introducing the terms of variational distribution and true posterior. We omitted the parameter identification $\theta$ and $\phi$ to save space in formula writing.
\mmyfont
\begin{equation}
\begin{split}
&{\rm log}\,p(R|C) \\
&=\int_{Z_\alpha}\int_{Z_p}q(Z_\alpha|P,R)q(Z_p|P){\rm log}\,p(R|C)dZ_p dZ_\alpha \\ \\
&=\int_{Z_\alpha}\int_{Z_p}q(Z_\alpha|P,R)q(Z_p|P) \\
&{\rm log}\,\frac{p(R|C)p(Z_p|C,R)q(Z_p|P)p(Z_\alpha|C,R,Z_p)q(Z_\alpha|P,R)}{p(Z_p|C,R)q(Z_p|P)p(Z_\alpha|C,R,Z_p)q(Z_\alpha|P,R)}dZ_p dZ_\alpha \\ \\
&=\int_{Z_\alpha}\int_{Z_p}q(Z_\alpha|P,R)q(Z_p|P) \\
&{\rm log}\,\frac{p(R|C)p(Z_p|C,R)p(Z_\alpha|C,R,Z_p)q(Z_\alpha|P,R)}{q(Z_p|P)p(Z_\alpha|C,R,Z_p)q(Z_\alpha|P,R)}dZ_p dZ_\alpha \\
&+\int_{Z_\alpha}\int_{Z_p}q(Z_\alpha|P,R)q(Z_p|P) {\rm log}\,\frac{q(Z_p|P)}{p(Z_p|C,R)}dZ_p dZ_\alpha \\ \\
&=\int_{Z_\alpha}\int_{Z_p}q(Z_\alpha|P,R)q(Z_p|P) \\
&{\rm log}\,\frac{p(R,Z_p|C)p(Z_\alpha|C,R,Z_p)q(Z_\alpha|P,R)}{q(Z_p|P)p(Z_\alpha|C,R,Z_p)q(Z_\alpha|P,R)}dZ_p dZ_\alpha \\
&+\int_{Z_p}q(Z_p|P) {\rm log}\,\frac{q(Z_p|P)}{p(Z_p|C,R)}dZ_p \\
\end{split}
\nonumber
\end{equation}
\normalsize
The first term can be factorized into two parts. We assume the true posterior $p(Z_\alpha|C,R,Z_p)$ is independent of the integrals over $Z_p$. Thus, the formula can be re-written as follows:
\mmyfont
\begin{equation}
\begin{split}
&{\rm log}\,p(R|C) \\
&=\int_{Z_\alpha}\int_{Z_p}q(Z_\alpha|P,R)q(Z_p|P){\rm log}\,\frac{q(Z_\alpha|P,R)}{p(Z_\alpha|C,R,Z_p)}dZ_p dZ_\alpha \\
&+\int_{Z_p}q(Z_p|P) {\rm log}\,\frac{q(Z_p|P)}{p(Z_p|C,R)}dZ_p \\
&+\iint q(Z_\alpha|P,R)q(Z_p|P){\rm log}\,\frac{p(R,Z_p|C)p(Z_\alpha|C,R,Z_p)}{q(Z_p|P)q(Z_\alpha|P,R)}dZ_p dZ_\alpha \\ \\
&\approx \underbrace{\int_{Z_\alpha}q(Z_\alpha|P, R){\rm log}\,\frac{q(Z_\alpha|P,R)}{p(Z_\alpha|C,R,Z_p)}dZ_\alpha}_{KL(q(Z_\alpha|P,R)||p(Z_\alpha|C,R,Z_p))} \\
&+\underbrace{\int_{Z_p}q(Z_p|P){\rm log}\,\frac{q(Z_p|P)}{p(Z_p|C,R)}dZ_p}_{KL(q(Z_p|P)||p(Z_p|C,R))} \\
&+\underbrace{\iint q(Z_\alpha|P,R)q(Z_p|P){\rm log}\,\frac{p(R,Z_p|C)p(Z_\alpha|C,R,Z_p)}{q(Z_p|P)q(Z_\alpha|P,R)}dZ_p dZ_\alpha}_{{\rm ELBO}}
\end{split}
\nonumber
\end{equation}
\normalsize
Where the first two terms are KL divergence between the true posterior and variational distribution. Since KL divergence is always greater than or equal to 0, to maximize the likelihood ${\rm log}\,p(R|C)$ can be converted to maximize ELBO, which can be reformulated as follows:
\mmyfont
\begin{equation}
\begin{split}
&{\rm log}\,p(R|C) \ge {\rm ELBO} = \\
&-\int_{Z_\alpha}\int_{Z_p}q(Z_\alpha|P,R)q(Z_p|P){\rm log}\,\frac{q(Z_p|P)q(Z_\alpha|P,R)}{p(R,Z_p,Z_\alpha|C)}dZ_p dZ_\alpha \\ \\
&=-\int_{Z_\alpha}\int_{Z_p}q(Z_\alpha|P,R)q(Z_p|P) \\
&{\rm log}\,\frac{q(Z_p|P)q(Z_\alpha|P,R)}{p(Z_p|C)p(Z_\alpha|C,Z_p)p(R|C,Z_p,Z_\alpha)}dZ_p dZ_\alpha \quad ({\rm Bayes' theorem}) \\ \\
&=\int_{Z_\alpha}\int_{Z_p}q(Z_\alpha|P,R)q(Z_p|P)p(R|C,Z_p,Z_\alpha)dZ_p dZ_\alpha \\
&-\int_{Z_\alpha}\int_{Z_p}q(Z_\alpha|P,R)q(Z_p|P){\rm log}\,\frac{q(Z_p|P)}{p(Z_p|C)}dZ_p dZ_\alpha \\
&-\int_{Z_\alpha}\int_{Z_p}q(Z_\alpha|P,R)q(Z_p|P){\rm log}\,\frac{q(Z_\alpha|P,R)}{p(Z_\alpha|C,Z_p)}dZ_p dZ_\alpha \\ \\
&\approx \underbrace{\iint q(Z_\alpha|P,R)q(Z_p|P)p(R|C,Z_p,Z_\alpha)dZ_p dZ_\alpha}_{\mathbb{E}_{q(Z_p|P);q(Z_\alpha|P, R)}[{\rm log}\,p(R|C, Z_p,Z_\alpha)]} \\
&-\underbrace{\int q(Z_p|P){\rm log}\,\frac{q(Z_p|P)}{p(Z_p|C)}dZ_p}_{KL(q(Z_p|P)||p(Z_p|C))} - \underbrace{\int q(Z_\alpha|P,R){\rm log}\,\frac{q(Z_\alpha|P,R)}{p(Z_\alpha|C,Z_p)}dZ_\alpha}_{KL(q(Z_\alpha|P,R)||p(Z_\alpha|C,Z_p))}
\end{split}
\nonumber
\end{equation}
\normalsize
We assume that the prior distribution $p(Z_\alpha|C,Z_p)$ is independent of the integrals over $Z_p$.
\end{document}